\pdfoutput=1

\documentclass[11pt]{article}

\usepackage{EMNLP2022}

\usepackage{times}
\usepackage{latexsym}

\usepackage[T1]{fontenc}

\usepackage[utf8]{inputenc}

\usepackage{microtype}

\usepackage{inconsolata}

\newcommand\blfootnote[1]{
  \begingroup
  \renewcommand\thefootnote{}\footnote{#1}
  \addtocounter{footnote}{-1}
  \endgroup
}

\usepackage{graphicx}
\usepackage{subfigure}
\usepackage{stmaryrd}
\usepackage{booktabs} %

\usepackage{amsmath}
\usepackage{amssymb}
\usepackage{mathtools}
\usepackage{amsthm}

\usepackage{cuted}
\usepackage{enumitem}
\usepackage{bm}
\usepackage{rotating}
\usepackage{multirow}
\usepackage[percent]{overpic}
\usepackage{listings}

\usepackage{hyperref}
\usepackage{url}
\usepackage{booktabs}
\usepackage{amsfonts}
\usepackage{nicefrac}
\usepackage{xcolor}

\title{Searching for Better Database Queries in the Outputs of Semantic Parsers}
  
\author{Anton Osokin\thanks{* Equal contribution.} \\
  HSE University \\
  Yandex\\\And
  Irina Saparina\footnotemark[1] \\
  HSE University \\
  Yandex\\\And
  Ramil Yarullin\footnotemark[1] \\
  HSE University \\
  Yandex\\}

\begin{document}
\maketitle
\blfootnote{Correspondence to: ramild.yar@gmail.com}
\begin{abstract}
The task of generating a database query from a question in natural language suffers from ambiguity and insufficiently precise description of the goal.
The problem is amplified when the system needs to generalize to databases unseen at training.
In this paper, we consider the case when, at the test time, the system has access to an external criterion that evaluates the generated queries. The criterion can vary from checking that a query executes without errors to verifying the query on a set of tests.
In this setting, we augment neural autoregressive models with a search algorithm that looks for a query satisfying the criterion.
We apply our approach to the state-of-the-art semantic parsers and report that it allows finding many queries passing all the tests on different datasets.

\end{abstract}
\section{Introduction}
Generating a database query from a natural-language description of the user's intent is a long-standing and important task.
In the recent years, most of the community focus was on the Spider dataset~\cite{yu-etal-2018-spider}, which poses the task in the zero-shot regime, meaning that a method has to generalize to databases unseen at training. 
The Spider dataset contains English questions and SQL queries.
The progress has been remarkable, and the accuracy has moved from below 30\% to above 70\%.
A part of this success can be attributed to the adoption of pre-trained transformer models like BERT~\cite{devlin-etal-2019-bert} into most of the pipelines. 

Given such progress, it is natural to ask whether we are getting closer to solving the problem.
Several recent studies have noted that the task might be harder than it looks.
\citet{finegan-dollak-etal-2018-improving} found that many single-database datasets had identical queries in both train and test sets and showed that using such splits effectively reduced the problem to classifying the queries from the train set. 
\citet{shaw-etal-2021-compositional} continued the study in the multi-database setting and showed that the compositional generalization was hard to achieve, and even to measure it, one should be very careful with splits.
In a different line of thought, \citet{suhr-etal-2020-exploring} examined how the models trained on Spider generalize to other datasets and reported that generalization was challenging. %
Even within one dataset, many questions have several interpretations leading to different queries, and annotation policies do not cover these ambiguities or cover them differently.

Acknowledging that the zero-shot setting might be too difficult to tackle as is, we aim to better define and simplify the problem to achieve better results in terms of the number of correctly generated queries.
In this task, most modern models produce a distribution over all possible outputs, which can guide the search at the test time.

We observe that if the search algorithm has access to a criterion that can evaluate the output by treating it as a database query, the overall method can produce much better results.
We consider the following criteria ordered by their ``strength'': a query is executed without errors, a query produces output from correct columns, a query produces a correct result on one test database, and a query produces correct results on a set of test databases.
We experiment with different search methods and report that the complete anytime beam search  \cite{zhang1998cab} outperforms sampling-based alternatives.

Many practical cases arise when the user is willing to trade off some of their time to improve the output query.
Our approach allows the user to obtain a better query by interactively guiding the search via providing the target output columns or the answer on one or more test databases.
The user is expected to supply this information without the gold query.
Notably, the execution criterion does not require extra user input but relies on executing generated queries. 

In addition, the test time database can be out of domain w.r.t.\ the training set. One can annotate gold queries for fine-tuning on new databases to improve out-of-domain performance, but this can be prohibitively expensive.
Our approach gives a way to improve out-of-domain results without extra annotated training data.
We view our approach as a way to control the trained model leading to a more reliable and responsible query synthesis.

For our studies, we used three state-of-the-art models in terms of execution accuracy (with publicly available implementations): T5-3B \cite{raffel-etal-2020-t5} fine-tuned to Spider by~\citet{scholak-etal-2021-picard}, BRIDGE \citep{lin-etal-2020-bridging} and
SQ-QDMR \citep{saparina-osokin-2021-sparqling}.

In this paper, we make several observations.
The complete anytime beam search  works with different search criteria and with all the three models.
Reasonable results can be obtained with the maximum beam size of 100, which fits on a single modern GPU.
 Searching with the execution criterion can significantly improve the quality of the decoders that generate output as an unconstrained token sequence, e.g., T5, and using such criterion does not require extra user input.
Searching for the queries that return a correct output on one database allows finding many queries that provide a correct output on that database.
Therefore, such searching w.r.t.\ one database can result in false positives, and it is important to evaluate the queries on a set of databases.
Based on the method of \citet{zhong-etal-2020-semantic}, we built the test suite of databases for the evaluation.
With these test suites, we show that there are multiple false positives among the queries that pass one test, while searching for the queries that pass the test suite produces the outputs of higher
quality.

Finally, we experiment with the generalization of the models trained on Spider to the GeoQuery~\cite{zelle1996learning}, IMDB, Yelp \citep{yelp_imdb} and
Academic \citep{academic} datasets.
We show that searching w.r.t.\ different criteria still works under this distribution shift, and searching w.r.t.\ the criteria with tests is often comparable to fine-tuning the network to a test dataset directly.

This paper is organized as follows.
In Section~\ref{sec:method_outline}, we review our setting.
In Section~\ref{sec:search_models}, we provide the details of our method.
Section~\ref{sec:test_suite} provides the details of the test-suite construction procedure, Section~\ref{sec:experimental_setup} describes the experimental setup.
In Section~\ref{sec:results}, we provide the experimental results and discussion.
We review some related works in Section~\ref{sec:related_works} and
conclude in Section~\ref{sec:conclusion}.

\section{Preliminaries \label{sec:method_outline}}
We consider the problem of generating queries to databases given the description of the user's intent in natural language in the cross-database setting where the train, validation and test splits contain different databases. Models trained in this setting, in theory, can be evaluated on any database.

A typical model for the cross-database setting is an encoder-decoder neural network.
Encoders typically consist of a pre-trained BERT-like transformer followed by a specialized encoder that can incorporate the database structure in some form \cite{guo-etal-2019-towards,wang-etal-2020-rat,cao-etal-2021-lgesql,cai2021}.
Sometimes the BERT part is further fine-tuned on database-related objectives 
\cite{yu2021grappa, deng-etal-2021-structure}.
The encoder input is a concatenation of the tokenized question and a sentence representation of the database schema separated by the special delimiter token. The representation of the database schema consists of the tokenized table and column names and values related to the question.
These values are commonly extracted by string matching with question tokens \cite{lin-etal-2020-bridging}.
Decoders are typically autoregressive based on LSTM or transformers.
Some decoders do not check the syntactic correctness of the output and its consistency with the database.
Some provide output w.r.t.\ a grammar~ \citep{yin-neubig-2017-syntactic}; some use post-hoc checks with parsers.

In this paper, we experiment with three models: T5-3B fine-tuned on the Spider dataset by \citet{scholak-etal-2021-picard}, BRIDGE \citep{lin-etal-2020-bridging} and
SQ-QDMR \citep{saparina-osokin-2021-sparqling}. %
We provide a detailed description of these models in Section~\ref{sec:models}.

\section{Search with Models \label{sec:search_models}}
We now describe our approach to searching for queries on top of a learned model.
We first generate full query candidates using a search method and then select the first one that passes the selected search criterion. We show possible search criteria in Section~\ref{sec:search_criteria} and search methods in Section~\ref{sec:search_methods}. 

\subsection{Search Criteria \label{sec:search_criteria}}

\paragraph{Execution criterion.}
To avoid syntactically incorrect queries, we can prune the search with the execution criterion. The query passes this criterion if it can be executed on the input database without errors. In particular, the query has to contain valid table and column names. These properties are not guaranteed for the unconstrained decoders as T5.
Thus the execution criterion can be extremely useful for such models.

\paragraph{Output column match.}
With this criterion, we compare the output columns (that the query will select) with the correct ones.
Firstly, wrong output columns is a common mistake in the text-to-SQL parsers \citep{guo-etal-2019-towards,lin-etal-2020-bridging, suhr-etal-2020-exploring}.
Secondly, the output columns can provide domain knowledge and shed some light on the user intent in a realistic scenario when the input question is ambiguous \citep{suhr-etal-2020-exploring, lee-etal-2021-kaggledbqa}.

\paragraph{One test.}
This criterion compares the result of query execution on a given database (one test case) with the correct one. 
With this criterion, we search for a correct query in terms of execution accuracy on the input database (Section~\ref{sec:exec_acc}).

\paragraph{Test suite.} This criterion checks if a query passes a set of tests.
Each test case corresponds to a particular database, and all test databases share the same schema.
This criterion is inspired by a test suite of databases with high code coverage proposed by \citet{zhong-etal-2020-semantic}.
The set of databases is designed to distinguish the correct queries from potential false positives. Searching with this criterion is equivalent to the search for a correct query in terms of test-suite accuracy (Section~\ref{sec:ts_acc}) .

\subsection{Search Methods \label{sec:search_methods}}

\paragraph{Top-$k$ and Top-$p$ (Nucleus) Sampling} \citep{fan-etal-2018-hierarchical,top-p-Holtzman2020} draw samples from the truncated distribution: the probability mass is re-weighted between the $k$ most probable elements in top-$k$ sampling and between the elements with cumulative probability mass exceeding $p$ in top-$p$ sampling ($k$, $p$ are hyperparameters).

\paragraph{UniqueRandomizer} \citep{unique-randomizer-shi-2020} is a method to incrementally sample sequences without replacement. The samples are drawn until the stopping condition is reached (one of the search criteria in our case). The probabilities of selected elements are reduced after each iteration of sampling to improve diversity in samples.

\paragraph{Complete Anytime Beam (CAB) Search}
of \citet{zhang1998cab} extends the regular beam search by running it several times with increasing beam sizes.
Importantly, the beams produced by beam search are known to have little diversity because of the peaks in the softmax scores.
We follow the approach of \citet{zohar2018,shrivastava2021}, who recently have used CAB to search for programs on top of neural autoregressive models.
In these works, the authors limit the number of hypotheses coming from each element of the previous beam (we will refer to this upper bound as the width of the beam search).
Between outer CAB iterations, we also increase the width by a constant value and multiply the beam size by a constant factor.
The schedules of the beam size and beam-search width are important hyperparameters.

\begin{table}[t]
\centering
\caption{Comparison of the test suites' statistics. \textbf{NoEmpty} is the percentage of SQL queries for which at least one test database with non-empty execution result is found; \textbf{Cover} is the percentage of neighbor queries distinguished by the test databases; \textbf{Tests} is the average number of test databases per query; \textbf{Time} is the average wall-clock execution time per query; \textbf{Size} is the total size of all test databases. \label{tab:ts_statistics}}
\begin{center}
\vskip -2mm
\begin{small}
\begin{tabular}{@{}l@{\;\:}l@{\!\!\!\!}c@{\;\:}c@{\;\:}c@{\;\:}c@{\;\:}c@{}}
\toprule
\multirow{2}{*}{\textbf{Dataset}} &  & \textbf{NoEmpty} & \textbf{Cover} & \textbf{Tests}  & \textbf{Time} & \textbf{Size} \\
 & & \textbf{(\%)} $\uparrow$ & \textbf{(\%)} $\uparrow$ & \textbf{(Num)} & \textbf{~~(Sec)} $\downarrow$ & \textbf{~~(GB)} $\downarrow$ \\
\midrule
\multirow{3}{*}{Spider} 
& 1 test & 96.7 & 89.1 & 1 & \textbf{0.1} & 3.25 \\ 
& Orig. & 98.2 & 96.1 & 675 & 3.6 & 0.16\\
& Our & \textbf{99.3} & \textbf{98.8} & 1.8 & 0.2 & \textbf{0.04}\\
\midrule
\multirow{3}{*}{GeoQuery} 
& 1 test & 94.7 & 93.2 & 1 & \textbf{0.1} &  $\bm{10^{\textbf{--4}}}$ \\ %
& Orig. & 66.7 & 52.6 & 108 & 12.6 & 0.01 \\
& Our & \textbf{100} & \textbf{98.9} & 1.6 & 0.2 & 0.05 \\
\midrule
\multirow{3}{*}{IMDB} 
& 1 test & 75.2 & 14.9 & 1 & 8.7  & 1.49\\ 
& Orig. & 79.2 & 79.4 & 200 & 17.7 & 0.04 \\
& Our & \textbf{100} & \textbf{99.6} & 2 & \textbf{0.3} & \textbf{0.03} \\
\midrule
\multirow{3}{*}{Yelp} 
& 1 test & 36.5 & 15.8 & 1 & 6.4 & 2.21\\
& Orig. & 84.1 & 80.6 & 282 &  30.9 & \textbf{0.03} \\
& Our & \textbf{99.2} & \textbf{97.0} & 3.1 & \textbf{0.4} & \textbf{0.03} \\
\midrule
\multirow{3}{*}{Academic} 
& 1 test & 51.1 & 5.50 & 1 & 44.2 & 4.33\\ 
& Orig. & \textbf{97.9} & 92.1 & 411 & 44.3 & \textbf{0.07} \\
& Our & 96.8 & \textbf{95.2} & 2.3 & \textbf{0.4} & \textbf{0.07} \\
\bottomrule
\end{tabular}
\end{small}
\end{center}
\vspace{-4mm}
\end{table}

\section{Test-Suite Construction \label{sec:test_suite}}
Testing on one database is generally not enough to ensure the semantic correctness of the generated query, but running the query on too many databases can be computationally inefficient. The inefficiency problem is especially acute in our task due to the large number of query candidates that should be tested and  several rounds of the searching process.

We build our test suites by modifying the method of \citet{zhong-etal-2020-semantic}, which relies on generating the so-called neighbor queries from a given set of gold queries and randomly sampling databases to distinguish gold queries from as many neighbors as possible.
We observed two key drawbacks of the test databases generated by \citet{zhong-etal-2020-semantic}. 
First, the test suites contained many databases, and some were unnecessarily large, which resulted in very long testing on them.
Second, outputs of many queries were often empty (or zero for queries with aggregators) on these test databases. If the output of the gold query is empty on all the elements of the test suite, it cannot be distinguished from trivial dummy queries.
This effect is more salient if the gold query returns empty output on the original database.
We alleviate these issues by independently generating test databases for each gold query, explicitly limiting the number of rows in each table and putting extra effort into generating at least one database where the gold query returns a non-empty output.
The details of our procedure are provided in Appendix~\ref{sec:app:testsutie}.

We compare our test suites with the original ones of \citet{zhong-etal-2020-semantic} on the five considered datasets described in Section \ref{sec:datasets}. %
For a fair comparison, we generate the independent sets of neighbor queries for each gold query. These neighbor queries are different from the neighbors generated in the process of creating the test suites. 
In Table~\ref{tab:ts_statistics}, we compare the initial databases (1 test), original test suites and our test suites. It can be seen that, in the space of neighbor queries, our test suites have higher code coverage (Cover) than the original databases and the original test suites. With new test suites, the average query execution time (on all the corresponding tests) is reduced 50x across the datasets.
We will release the test suites we created.

\section{Experiment Setup \label{sec:experimental_setup}}
\subsection{Data \label{sec:datasets}}

We conduct our experiments on five text-to-SQL datasets: multi-database Spider \citep{yu-etal-2018-spider} and single-database GeoQuery \cite{zelle1996learning,iyer-etal-2017-learning,finegan-dollak-etal-2018-improving}, IMDB, Yelp \citep{yelp_imdb} and
Academic \citep{academic}. 

\paragraph{Spider.} We use dev and test sets (451 and 521 examples) from the work of \citet{saparina-osokin-2021-sparqling}: they are parts of the original Spider dev, but some examples (from dev) were repaired.

\paragraph{GeoQuery, IMDB, Yelp and Academic.}
We use query splits created by \citet{finegan-dollak-etal-2018-improving}, additionally filtered from duplicates and examples with gold SQL queries that crash or execute longer than 5 minutes with the Python package sqlite3.
The dataset statistics are provided in Table~\ref{tab:datasets_stats} of Appendix~\ref{apn:datasets}.

We do not consider more single-database datasets because ATIS \citep{atis1, atis2}, Scholar \citep{iyer-etal-2017-learning} and Advising \citep{finegan-dollak-etal-2018-improving} database schemes exceed 512 token limits of pre-trained encoders, Restaurants \citep{restaurants1, restaurants2} contains too many duplicates.

\subsection{Evaluation Metrics  \label{sec:exec_acc}} 
\paragraph{Exact-set Match} \citep{yu-etal-2018-spider} is an SQL-to-SQL comparison metric that reflects the fraction of the predicted queries matching the ground-truth queries. In the matching process, each query is decomposed into fragments that are compared individually so that the metric is not too sensitive to the ordering of independent clauses. This metric does not take into account predicted values and can give a high score to incomplete queries. As SQ-QDMR model produces queries in SPARQL, we cannot use the exact-set match as a primary evaluation metric. 

\paragraph{Execution accuracy} is designed to compare the queries by their execution output on an original database. In contrast to the exact-set match, this prevents false-negative queries but leaves space for potential false positives. The version provided by \citet{yu-etal-2018-spider} for Spider evaluation has issues in SPARQL-SQL comparison, so we use the version provided by \citet{saparina-osokin-2021-sparqling} %
 unless explicitly mentioned otherwise. 

\paragraph{Test-suite accuracy\label{sec:ts_acc}} \citep{zhong-etal-2020-semantic} approximates the semantic accuracy of the query synthesis models. This metric refers to the share of predicted queries producing the correct answers on all databases from the test suite. We build the test suites for Spider dev and test sets and for all the queries in the other four considered datasets. 

\subsection{Models \label{sec:models}}
We consider three models: T5-3B fine-tuned on Spider \citep{scholak-etal-2021-picard}, BRIDGE \citep{lin-etal-2020-bridging} and SQ-QDMR \citep{saparina-osokin-2021-sparqling}.
These models have top execution accuracy among publicly available models on Spider.
We also tried to search under our search criteria on top of the bottom-up semi-autoregressive model of \citet{rubin-berant-2021-smbop-semi}, but we could not make the search increase the number of correct queries.

For evaluation on Spider, we use the released checkpoints of the best models.
BRIDGE training data included question splits of single-database datasets, so we re-train it on Spider-only data to evaluate on query splits of these datasets.
For re-training BRIDGE and fine-tuning all models, we use official implementations (see Appendix~\ref{apn:finetune_params}). 

\paragraph{T5} \citep{raffel-etal-2020-t5} is a pre-trained seq2seq model based on Transformer. 
Recently, \citet{shaw-etal-2021-compositional, scholak-etal-2021-picard} successfully applied T5 for the text-to-SQL task. 
The input sequence contains question tokens and tokens of column and table names. The database values matched with the question tokens are appended to the corresponding column names \citep{lin-etal-2020-bridging}.
The output of the T5 model is the sequence of tokens representing the SQL query. Note that this model generates output sequence without explicitly considering the SQL grammar and schema consistency.

\paragraph{BRIDGE} \citep{lin-etal-2020-bridging} consists of the BERT-based encoder and pointer-generator decoder. The input sequence is formed from the concatenation of question, table and column names, and relevant database values separated by special token and encoded with BERT. The relevant values are selected with fuzzy string matching between question and database values. Column encodings are further enriched with meta-data features such as primary or foreign keys and data types obtained from the feed-forward layer. 
The LSTM-based decoder with multi-head attention at each step copies question or schema tokens or generates the SQL keywords. During decoding, model chooses columns only from the predicted table %
to provide schema consistency. An additional static SQL analyzer filters incorrect output queries.

\paragraph{SQ-QDMR} (as we refer to the model of \citet{saparina-osokin-2021-sparqling}) contains RAT-transformer, GraPPa encoders \cite{wang-etal-2020-rat,yu2021grappa} and a grammar-guided LSTM-based decoder \cite{yin-neubig-2017-syntactic}.
The  SQ-QDMR decoder produces output in the form of grounded intermediate representations derived from QDMRs of the Break dataset \cite{wolfson-etal-2020-break}.
The grounded QDMRs are not directly related to any execution engine and cannot be executed as is, but \citet{saparina-osokin-2021-sparqling} implemented a non-trainable translator from grounded QDMR to the SPARQL query language, in which queries can be executed. %
We can think of grounded QDMRs augmented with this translator as executable database queries.

\section{Results \& Analysis \label{sec:results}}
\subsection{Impact of the Search Methods}
We compare different decoding strategies in our setting (Table~\ref{tab:samplings}): top-$k$ and top-$p$ (nucleus) sampling \citep{fan-etal-2018-hierarchical,top-p-Holtzman2020}, UniqueRandomizer \citep{unique-randomizer-shi-2020} and CAB search \cite{zhang1998cab}.
We measure the execution accuracy of these search methods under the 1-test criterion on Spider dev. We use the same sampling budgets (1000 for BRIDGE and SQ-QDMR and 800 for T5-3B due to the memory limits), tune $p$ in top-$p$ sampling and the temperature for all methods, more implementation details in Appendix~\ref{apn:search_details}).

The results demonstrate that a significant number of output queries pass one test after searching with any of these methods, so different decoding strategies can be compatible with our approach. 
UniqueRandomizer is very time-consuming since it generates samples sequentially in contrast to other methods that generate beams of samples in parallel.
CAB search is demanding in terms of the device memory as it has to process the whole beam jointly.
For further experiments, we choose CAB search because it works best for two models.

\subsection{Impact of the Search Criteria}
We apply search under different selection criteria (execution, output column match, test on one database) to T5-3B, BRIDGE and SQ-QDMR on Spider dataset and compare with the greedy and beam search baselines.
Table~\ref{tab:search_spider_exec} shows the results measured with execution accuracy.
\textbf{Searching on top of all models with different selection criteria increases execution accuracy in almost all cases.}

\begin{table}[t]
\caption{Comparing the execution accuracy of different search approaches under the 1-test criterion on the Spider dev split.
\label{tab:samplings}}
\begin{center}
\vskip -2mm
{\small
\begin{tabular}{@{}l@{\;\;}c@{\;\;}c@{\;\;}c@{\;\;}c@{\;\;}c@{}}
\toprule %
\textbf{Model} & \textbf{Greedy} & \textbf{CAB}  & \textbf{Sample} & \textbf{Top-p} &  \textbf{UniRand} \\
\midrule %
T5-3B &    77.0 &  \textbf{94.4} & 93.0 & 93.0 & 94.1 \\ %
BRIDGE &   66.8 & \textbf{91.0} &  87.1 & 84.9 & 86.2 \\ %
SQ-QDMR &   80.4 & 98.0 & 98.0 & 98.0 & \textbf{98.2} \\ %
\bottomrule
\end{tabular}
}
\end{center}
\vspace{-4mm}
\end{table}

One exception is the search with the execution criterion on top of BRIDGE and SQ-QDMR, the results of which are close to the greedy decoding.
The outputs of these systems are almost always executable because BRIDGE runs a static SQL analyzer for filtering, SQ-QDMR decodes according to the QDMR grammar  %
and both models have schema-consistent decoding. 
The T5 model, in contrast, does not have any grammar or schema constraints in decoding. \textbf{For language model decoders that do not explicitly check grammar and schema consistency, the execution criterion can significantly improve the quality.}

For T5-3B, we also compare the results of the search with PICARD \citep{scholak-etal-2021-picard}, our search with the execution criterion, %
and greedy decoding.
For a fair comparison with \citet{scholak-etal-2021-picard}, we use the same data, the official Spider dev set, and the same metrics: exact-set matching accuracy and execution accuracy provided by \citet{yu-etal-2018-spider}:

\begin{center}
\small 
\begin{tabular}{@{}lcc@{}}
\toprule %
\textbf{Method}  & \textbf{EM} & \textbf{Exec}   \\
\midrule %
Greedy & 71.5 & 74.4 \\ 
PICARD & 75.5 & 79.3 \\
CAB+execution    & 74.5  & 78.7 \\
\bottomrule
\end{tabular}
\end{center}

The results obtained with PICARD and with searching under the execution criterion are comparable~-- our search gets most of the benefit over the baseline.
PICARD provides slightly better quality and works with smaller beams but requires more effort to incorporate because it is tightly connected with the decoder
output vocabulary and grammar.
For both approaches, the percent of output queries with execution errors is around 2\% in contrast to the baseline T5-3B decoding with 12\%.

\textbf{The search criterion based on the matching of output columns provides even better results.}
As Table~\ref{tab:search_spider_exec} shows, all models benefit from this criterion: execution accuracy increases by 3-4\% nearly everywhere.
This criterion largely simplifies the task  with extra information at the test time.
\begin{table}[t]
\caption{Execution accuracy of search under different selection criteria (execution, output column match and 1 test) on Spider; beam s. refers to beam search. \label{tab:search_spider_exec}}
\begin{center}
\vskip -2mm
\small
\begin{tabular}{@{}l@{\;\;\;}c@{\;\;\;}c@{\;\;\;}c@{\;\;\;}c@{\;\;\;}c@{\;\;\;}c@{}}
\toprule %
\textbf{Model} & \textbf{Split}  & \textbf{Greedy} &  \textbf{Beam S.}  & \textbf{Exec} & \textbf{Cols} & \textbf{1 Test}  \\
\midrule %
T5-3B   &    \multirow{3}{*}{dev}                  & 77.0 & 77.4 & 83.1	& 84.2 & 94.4 \\
BRIDGE  &  & 66.8 & 68.8 & 68.4 & 72.0 & 91.0 \\
SQ-QDMR &                      & 80.4 & 80.6 & 80.4 & 83.6 & 98.0 \\
\midrule
T5-3B   &    \multirow{3}{*}{test}                   & 70.8 & 71.0 & 73.7 & 77.4 & 90.7 \\
BRIDGE  &  & 64.0 & 66.2 & 64.6 & 67.9 & 83.4 \\
SQ-QDMR &                       & 65.6 & 65.8 & 65.6 & 68.9 & 86.7 \\
\bottomrule
\end{tabular}
\end{center}
\vspace{-4mm}
\end{table} 

\textbf{Search for the queries that pass one test allows finding a significant number of such queries}. Passing one test means correct execution result on the input database, so all the queries found with this criterion are correct in terms of the execution accuracy.
Thus, these results indicate that our searching approach works, and we can find the correct queries with the corresponding criterion.
However, we observe that more than 10\% of queries found with one-test criterion are false-positive according to the test-suite accuracy (Table~\ref{tab:exec_ts} in Appendix~\ref{apn:exec-ts}).

These results motivate us to evaluate the test-suite accuracy of criterion-guided search.
Table~\ref{tab:search_spider_ts} confirms our findings: searching with the execution criterion helps T5-3B, and searching for correct output columns improves the results of all models. 
\textbf{Search for the queries that pass one test results in a significant number of false-positive queries. The correct queries can be found by searching with the test-suite criterion directly.}

\begin{table}[t]
\caption{Test-suite accuracy of search under different selection criteria (execution, output column match, 1~test and test suite) on Spider. \label{tab:search_spider_ts}}
\begin{center}
\vskip -2mm
\small
\begin{tabular}{@{}l@{\;\;\;}c@{\;\;}c@{\;\;}c@{\;\;}c@{\;\;}c@{\;\;}c@{}}
\toprule %
\textbf{Model} & \textbf{Split} & \textbf{Greedy} & \textbf{Exec} &  \textbf{Cols} &  \textbf{1 Test} & \textbf{Test Suite} \\
\midrule %
T5-3B   &    \multirow{3}{*}{dev}                  & 72.2 & 77.7 & 78.3 & 86.9 & 90.1 \\ %
BRIDGE  &  & 63.2 & 64.8 & 68.2 & 81.9 & 84.0 \\
SQ-QDMR &                      & 72.5 & 72.2 & 75.4 & 84.0 & 94.4 \\ %
\midrule
T5-3B   &    \multirow{3}{*}{test}                & 68.9 & 73.7 & 75.4 & 84.7 & 86.8 \\ %
BRIDGE  &   & 61.9 & 62.5 & 65.4 & 76.6 & 77.8 \\
SQ-QDMR &                    & 62.3 & 62.1 & 65.4 & 75.2 & 84.1 \\ %
\bottomrule %
\end{tabular}
\end{center}
\vspace{-4mm}
\end{table}

\subsection{Efficiency}
\paragraph{Time Measurements.}
The running time during the search is dominated by the time of the decoder for all three models:
executing each considered query takes 3\% of the decoder time for T5-3B, which is 0.01 sec per run; 53\%, 0.02 sec~-- for BRIDGE; 72\%, 0.03 sec ~-- for SQ-QDMR.
The total running time depends on the effective beam size used during the search.

The T5-3B model with the execution criterion on top runs in 1.7 sec compared to 3.1 sec reported by the PICARD paper, where both systems were run on 1 NVIDIA A100 GPU.
The main reason is that due to CAB, we do not set one beam size in advance and thus, process at least 70\% of examples with an effective beam size of 1.

\begin{figure}[t]
\vskip -1.25mm
\begin{center}
\centerline{\includegraphics[scale=0.7]{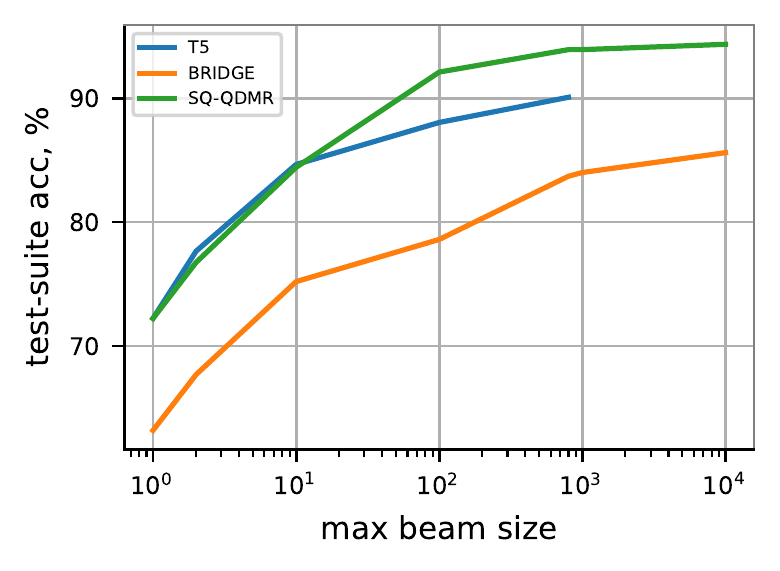}}
\vskip -2mm
\caption{Test-suite accuracy on the dev set for the T5-3B, BRIDGE and SQ-QDMR models tested with CAB for the test-suite criterion.\label{max-bs}}
\end{center}
\vspace{-6mm}
\end{figure}

\paragraph{Impact of the Maximum Size of Beam. \label{sec:max-beam}} 
The maximum size of the beam is an important parameter.
Figure~\ref{max-bs} shows the dependence between the obtained test-suite accuracy on the Spider dev set and the maximum beam size in the search under the test suite criterion.
For all models, we start with the maximum beam size equal to 1, which is equivalent to the greedy decoding and finish with the maximum beam size allowed by our implementation and hardware: 10k for BRIDGE and SQ-QDMR on 1 NVIDIA V100 GPU and 800 for T5-3B on 8 NVIDIA A100 GPUs. 

Test-suite accuracy improves as the maximum beam size increases.
BRIDGE with the maximum beam size equal to 10k achieves 86\% of test-suite accuracy and SQ-QDMR --- 94\%.
However, the search works well enough even with smaller beams: with the maximum beam size of 100, BRIDGE achieves almost 80\%, SQ-QDMR  --- 92\%, and T5-3B achieves 88\% (with 800, T5-3B achieves 90\%). Importantly, the search with the beam size of 100 does not require multiple GPUs for T5-3B.
\subsection{Experiments on Single-Database Data} %

To show more benefits of searching under selection criteria, we evaluate it on single-database datasets, GeoQuery, IMDB, Yelp and Academic, with test-suite accuracy (Table~\ref{tab:geo-testsuite}; see Appendix~\ref{apn:others-exec} for execution accuracy). We use query splits of \citet{finegan-dollak-etal-2018-improving} and two model types: trained on Spider only and fine-tuned on a particular dataset. The Academic database is very large, so we cannot evaluate one-test criterion on this dataset and fine-tune SQ-QDMR 
(other datasets do not have QDMR annotation required for fine-tuning).
More fine-tuning details are  in Appendix~\ref{apn:finetune_params}. %

The results show that models trained on Spider struggle to generalize to other datasets, which is consistent with the findings of \citet{suhr-etal-2020-exploring}.
\textbf{More information about the problem (in the form of additional train data or selection criteria) helps improve the quality.}

IMDB, Yelp and Academic are more challenging datasets for cross-database semantic parsers than GeoQuery, but they are significantly smaller (Table~\ref{tab:datasets_stats}), and the models are less stable while testing on them (with all random seeds fixed). %
Stronger criteria, such as passing one test or test suite, do work even with datasets of such difficulty, when weaker criteria fail.
On GeoQuery, the search under the one-test and test-suite criteria leads to even better quality than fine-tuning. Our test-suite criterion is especially useful when one test is difficult to run on the large original database, e.g., on Academic.

As a result, we conclude that criterion-guided search on top of a pre-trained model is a good alternative to fine-tuning in cases when training data is not available, but the user is ready to provide more information on each test question.

\begin{table}[t]
\centering
\caption{Different search criteria (execution, output column match, 1 test and test suite) on top of pre-trained models on the query test splits of different datasets with the \textbf{test-suite accuracy}. \label{tab:geo-testsuite}}
\begin{center}
\begin{small}
\vskip -2mm
\begin{tabular}{@{}c@{\;\;}l@{}c@{\;\;}c@{\;\;}c@{\;\;}c@{\;\;}c@{}}
\toprule
\multirow{2}{*}{\shortstack{\textbf{Dataset} \\ \textbf{(test size)}}} & \multirow{2}{*}{\textbf{Model}} & \multirow{2}{*}{\textbf{Greedy}} & \multirow{2}{*}{\textbf{Exec}} & \multirow{2}{*}{\textbf{Cols}} & \multirow{2}{*}{\textbf{1 test}} & \multirow{2}{*}{\shortstack{\textbf{Test} \\ \textbf{Suite}}} \\
 &  & &  &  &  &  \\
\midrule
\multirow{6}{*}{\shortstack{GeoQuery \\ (182)}} & T5-3B   &  55.5  & 56.6 & 63.7 & 67.6 & 72.5 \\ 
& + fine-tune & 64.3 & 70.9 & 85.2& 88.5 & 94.5 \\[0.7ex]
& BRIDGE & 55.9 & 50 & 62.6 & 74.7 & 74.7 \\
& + fine-tune & 65.4 & 66.5 & 81.9 & 86.8 & 91.8 \\[0.7ex]
& SQ-QDMR & 37.4 & 37.4 & 41.8 & 65.9 & 81.9 \\
&  + fine-tune &  56.0  & 56.0 & 59.9 & 76.9 & 83.0 \\
\midrule
\multirow{5}{*}{\shortstack{IMDB \\ (17)}} & T5-3B   & 5.9 & 11.8  & 17.6 & 29.4 & 41.2 \\ 
& + fine-tune & 52.9 & 52.9  & 52.9  & 52.9  & 58.8 \\[0.7ex]
& BRIDGE & 11.8 & 11.8 & 17.6 & 17.6 & 17.6 \\
& + fine-tune & 52.9  &52.9 & 52.9 & 52.9 & 52.9 \\[0.7ex]
& SQ-QDMR & 5.9 & 5.9 & 11.8 & 29.4 & 47.1 \\
\midrule
\multirow{5}{*}{\shortstack{Yelp \\ (10)}} & T5-3B & 20 & 20 & 10 & 30 & 10  \\ 
& + fine-tune& 20 & 30 & 20  & 50 &  40\\[0.7ex]
& BRIDGE & 0 & 0 & 0 &  20 & 10 \\
& + fine-tune & 30 & 40 &  50 &  60 & 70 \\[0.7ex]
& SQ-QDMR &  10 & 10 & 10 & 10 & 40   \\
\midrule
\multirow{5}{*}{\shortstack{Academic \\ (15)}} & T5-3B  & 6.7 & 13.3 & 26.7 & - %
& 33.3 \\ 
& + fine-tune & 60 & 53.3 & 53.3 & - %
&   80 \\[0.7ex]
& BRIDGE & 0 & 0 & 6.7  & - & 20 \\
& + fine-tune & 33 & 40 & 40 & - & 80 \\[0.7ex]
& SQ-QDMR &  13.3 & 13.3 & 6.7 & - %
&  66.7 \\
\bottomrule
\end{tabular}
\end{small}
\end{center}
\vspace{-4mm}
\end{table}

\section{Related Works \label{sec:related_works}}

\paragraph{Search for Database queries.}
The task of translating NL questions into database queries implies the ability to query databases with natural language. To ensure this, it is essential to generate syntactically correct queries that refer to valid table and column names for the given database schema. \citet{ExecutionGuided2018} noticed that a partially decoded SQL query can be executed, and thus, the result of this execution can guide the decoding process. At each decoding step, partial queries that crush or give an empty result during the execution are removed from beams. In this work, we also consider the execution criterion of search but apply it to the finished hypotheses, which allows us to search on top of the models with different output formats, including intermediate representations.

\citet{lin-etal-2020-bridging} generated SQL queries in the execution order to keep the consistency between the predicted database entities and checked output correctness with the static SQL parser.
\citet{suhr-etal-2020-exploring} executed the top-10 generated queries in beam search to filter the inexecutable ones, which is close to checking the execution criterion in our work but differs by the search method. 

Task-specific decoders such as autoregressive grammar-based  \citep{yin-neubig-2017-syntactic,GrammarBased2018} and tree decoders \citep{dong-lapata-2016-language}, semi-autoregressive  decoder \citep{rubin-berant-2021-smbop-semi} provide some guarantees as they control the output structure. However, as noticed by \citet{scholak-etal-2021-picard}, these decoding methods are incompatible with pre-trained decoders of language models.
These pre-trained decoders, like the one of T5, can also be successfully applied to the text-to-SQL task \citep{shaw-etal-2021-compositional}. \citet{scholak-etal-2021-picard} proposed to check hypotheses in beam search on the lexical and grammatical levels at each step of the beam search.
However, compared to us, their approach required heuristics to prune incomplete queries.

The concurrent work by \citet{wolfson-etal-2022-weakly} uses several components similar to ours but in a very different way.
They use the QDMRs of~\citet{wolfson-etal-2020-break} with textual arguments as a form of weak supervision to generate SQL queries for the training set.
Their synthesis process results in many candidate SQL queries and relies on tests to select the one as an annotation.
Such a process is similar to the method of~\citet{saparina-osokin-2021-sparqling} for constructing groundings of QDMR arguments.
However, the search process of \citet{wolfson-etal-2022-weakly} is not connected to any neural model and is not used at the test time.

\paragraph{Search for Programs with Neural Networks.}
Our approach to searching for queries is closely related to the field of program synthesis if we interpret queries as programs.
Recently, neural networks have been applied in a wide range of program synthesis tasks,
see the excellent work of \citet{chaudhuri2021tutorial} for a recent review.

When programs are synthesized from large language models generating multiple outputs, selecting the one that, e.g., passes some or all tests is a common practice.
For example, the Codex model~\cite{chen2021codex} for synthesizing Python code includes some sample tests into the input prompt to give the model more information to define the user intent.
\citet{chen2021codex} following \citet{kulal2019spoc}, among others, also uses the pass@k metric, which effectively means that the model generates $k$ outputs, and the best ones are selected based on tests.
The pass@k metric can be interpreted as test-suite accuracy after search w.r.t.\ the tests with the beam of size $k$.

Overall, it is widely accepted that tests are useful to precisely define the user intent.
However, they are hard to collect at a large scale, especially when coupled with a description in natural language.
Because of this, large-scale benchmarks related to code, e.g., CodeXGLUE \cite{lu2021codexglue}, primarily used text-based metrics like BLEU. 
The attempts to specialize BLEU to code by combining it with abstract syntax trees extracted from code, like CodeBLEU \cite{ren2020codebleu}, are in some sense similar to the SQL-based exact match metric of the Spider dataset~\cite{yu-etal-2018-spider}.

\paragraph{Approaches to Simplify the Cross-database Setting.}
The community has made multiple attempts to modify the cross-domain setting to make the problem easier to solve.
\citet{yu-etal-2019-sparc} collected the SParC dataset with coherent question sequences, which can allow sharing of information between the sequences.
\citet{yu-etal-2019-cosql} collected the CoSQL dataset with an interactive conversational setting with SQL queries, making it possible to explore user interactions with the system.
\citet{lee-etal-2021-kaggledbqa} collected KaggleDBQA with database documentation in the form of textual description for database columns that can potentially allow language models to provide outputs better corresponding to the user's intent.
Our approach provides users with a way to interact with the model by supplying tests.
Given the initial question in natural language, the users can provide extra tests until they are satisfied with the generated query.

\section{Conclusion \label{sec:conclusion}}

In this paper, we studied the search over the outputs of the neural autoregressive models for better database query generation. We considered three state-of-the-art models: T5-3B, BRIDGE and SQ-QDMR.
We observed that the search algorithms work with multiple criteria for selecting the output query.
We also compared the search-augmented methods with the fine-tuned models on the GeoQuery, IMDB, Yelp and Academic datasets (under the distribution shift) and observed that the method with search can sometimes work even better than fine-tuning.
Compared to fine-tuning, the search-based method does not require additional training data but relies on additional information on each test example.
With such properties, our search based methods might be helpful for use cases like interactive query generation or annotating new datasets.

\section*{Limitations}
In our experiments, we worked only with the datasets where the user question was written in English.
This might have simplified the task for T5 as the keywords and entity names of the query languages were also in English.

The test suites we built were still not perfect.
In particular, it was hard to generate a test database such that the query {\footnotesize\texttt{SELECT year FROM concert GROUP BY year HAVING count(*) >= 50}} had non-empty output because it needed at least 50 rows with the same value in the column {\footnotesize\texttt{year}}.
We also noticed that the value 1 as the gold query output also caused many false positives and should probably have been considered the empty value for the queries outputting the count aggregator.

The results of the search methods were also not perfect for out-of-domain data, even with strong search criteria based on tests.
One reason for that was that we started to hit the limitations of the models, which were built with mostly Spider in mind.
In particular, the database preprocessing stage to select values for the query was, in some cases, slow and inaccurate.

\bibliography{anthology,custom}
\bibliographystyle{acl_natbib}

\appendix

\section{Construction of Test Suites  \label{sec:app:testsutie}}

The method of \citet{zhong-etal-2020-semantic} relies on generating the so-called neighbor queries from a given set of gold queries $Q$. A set of neighbors $N_q$ is obtained through slightly changing the original query $q \in Q$. Now, a database $w$ distinguishes $q$ and $g \in N_q$ if their execution results on $w$ are different. The suite $S$ is formed greedily: a new sampled database is added to $S$ if it distinguishes a pair in $\mathcal{N} = \{(q, g) \mid q \in Q, g \in N_q\}$ that is not distinguished by any database previously added to~$S$.

In the algorithm above, to sample the databases, the original queries are parsed to derive the constant values and the corresponding columns from the \texttt{WHERE} clauses. For instance, in a query \texttt{SELECT * from cars WHERE mpg > 20}, the constants 20 and 21 are assigned some probability to be generated as values of the \texttt{cars.mpg} column. We improve the query parser so that it accounts for common cases when table aliases are used (e.g., \texttt{SELECT * from cars as T WHERE T.mpg > 20}). Still, if a gold query has too many \texttt{WHERE} clauses, its execution on a randomly generated database is likely to be an empty table or, in the case of \texttt{SELECT}-ing an aggregate function, zeros or \texttt{NULL} values. This issue causes many potential false positives.

For this reason, for every query, we first search for a test database on which the query execution is not empty and only then proceed to check if the database distinguishes the query and its neighbors. This way, the chances are higher to sample at least one database yielding non-empty output on a gold query, leading to better code coverage. We also propose to build a separate test suite for each query. This modification allows dramatically reducing the time of the test-suite evaluation of a query.

Additionally, we make several important changes to the original implementation:
\begin{itemize}[noitemsep,topsep=0mm,leftmargin=3mm] %
    \item We limit the number of rows in the tables to 100. As a result, the test suite databases are smaller in size and faster in evaluation.
    \item We adjust the types of the columns in the original and the sampled databases so that they match the type of the corresponding values. In the float type columns, we also reduce the precision of numbers to 16-bit. These changes are crucial for SPARQL language, as it is more sensitive to data types than SQL.
    \item If a column has only unique values in the original database and its name contains special substrings such as `Name', `ID', and `Phone', we treat it as a unique key and generate its values accordingly. This heuristic allows performing the \texttt{GROUP{\tiny~}BY} operations on any unique key.
\end{itemize}

\section{Implementation Details of Search \label{apn:search_details} }
For the search on top of T5-3B, we use the \citep{scholak-etal-2021-picard} model provided in the Transformers library \citep{wolf-etal-2020-transformers}. To search with large beam sizes, we modify the Transformers implementation of T5 inference: instead of caching keys and values in attention blocks, we cache the attention outputs, which reduces memory usage at the cost of speed.
 
For working with SQL queries outside the Spider dataset, we had to replace the SQL parsing coming with Spider due to its limited functionality.
We use the mo-sql-parsing library instead.\footnote{\href{https://github.com/klahnakoski/mo-sql-parsing}{github.com/klahnakoski/mo-sql-parsing}}
We execute all SQL queries in the sqlite3 \footnote{\href{https://docs.python.org/3/library/sqlite3.html}{docs.python.org/3/library/sqlite3.html}} package for python.
For executing SPARQL queries, we use the open-source version of the Virtuoso system.\footnote{\href{https://github.com/openlink/virtuoso-opensource}{github.com/openlink/virtuoso-opensource}}
During our search, the system generated many cumbersome queries, so we had to impose a strict time limit for each query and make our implementation robust to the database engine crashes. The time limit for Spider queries is 30 seconds and for other datasets is 300 seconds.

The grid of beam sizes for CAB search on top of T5-3B is 2, 10, 100, 800, the corresponding widths are 2, 2, 2, 5, the grid for BRIDGE is 1, 10, 100, 1000 and the widths are 1, 2, 2, 5, the grid for SQ-QDMR is 1, 100, 1000 and the widths are 1, 5, 10. We infer BRIDGE and SQ-QDMR on one NVIDIA V100 GPU and T5-3B on 8 NVIDIA A100 GPUs (while searching with maximum beam size). 
For the search with the sampling methods, we use the same schema: we run sampling several times, increasing the number of samples ($k$ in top-$k$) until the selection criterion is passed. The grids for all the methods are the same. For searching with UniqueRandomizer, we run the methods until the selection criterion is passed or the maximum number of iterations is reached (equal to the maximum beam size: 800 for T5-3B and 1000 for BRIDGE and SQ-QDMR).
We tune the temperature and $p$ on Spider dev (the grid for temperature tuning was from 0.5 to 3.5 with 0.25 step; the grid for $p$ was 0.85, 0.9, 0.95). The best $p =0.95$ and the best temperature values are the following:

\begin{center}
\small 
\begin{tabular}{lcccc}
\toprule %
\textbf{Model} & \textbf{CAB} & \textbf{Sample} & \textbf{Top-p} & \textbf{UniRand} \\
\midrule %
T5-3B   & 2.7 & 2.4 & 3.1 & 2.7 \\
BRIDGE  & 2.6 & 3.4 & 3.7 & 3.5 \\
SQ-QDMR & 1.0 & 1.3 & 1.9 & 1.2  \\
\bottomrule
\end{tabular}
\end{center}

While testing on single-database data, we use the same temperature parameters for the models trained on Spider only. For the fine-tuned models, we also tune temperature on dev sets and choose the values 2.0 for BRIDGE and T5-3B and 1.0 for SQ-QDMR. 

\section{Execution (1-test) Accuracy vs. Test-suite Accuracy \label{apn:exec-ts}}

In Table~\ref{tab:exec_ts}, we show the results of the search on Spider with two criteria: passing one test and passing the test suite.
For both criteria, we compute execution accuracy (checks the execution result on one database) and test-suite accuracy (checks the execution results on the test suite).
For all the models, test-suite accuracy is significantly lower than execution accuracy for the search with the one-test criterion. 

\begin{table}[ht]
\caption{Execution (EX) and test-suite (TS) accuracy of search with 1-test and test-suite criteria on Spider dev and test. \label{tab:exec_ts}}
\begin{center}
\vskip -2mm
{\small
\begin{tabular}{@{}l@{\;\;\;\;}c@{\;\;\;\;}c@{\;\;\;\;}c@{\;\;\;\;}c@{\;\;\;\;}c@{}}
\toprule \multirow{2}{*}{\textbf{Model}} &  \multirow{2}{*}{\textbf{Split}} 
 & \multicolumn{2}{c}{\textbf{1 test}}  & \multicolumn{2}{c}{\textbf{Test Suite}} \\
\cmidrule{3-6}
 &  & EX  & TS   & EX  & TS  \\
\midrule %
T5-3B   &  \multirow{3}{*}{dev}  & 94.4 & 86.9 & 91.4 & 90.1 \\
BRIDGE  & & 91.0 & 81.9 & 87.4 & 84.0 \\
SQ-QDMR &                      & 98.0 & 84.0 & 93.7 & 94.4 \\
\midrule
T5-3B  & \multirow{3}{*}{test} & 90.7 & 84.7 & 85.5 & 86.8 \\
BRIDGE   &   & 83.4 & 76.6 & 77.4 & 77.8 \\
SQ-QDMR &    & 86.7 & 75.2 & 79.7 & 84.1 \\
\bottomrule
\end{tabular}}
\end{center}
\vspace{-4mm}
\end{table}

\section{Single-database Datasets \label{apn:datasets}}
We use the query splits provided by \citet{finegan-dollak-etal-2018-improving} for four single-database datasets: GeoQuery, IMDB, Yelp and Academic. We choose query splits because they are more difficult than question splits, according to the findings of \citet{finegan-dollak-etal-2018-improving}. We exclude duplicates and examples with gold SQL queries that crash or execute longer than 5 minutes with sqlite3. The statistics of resulted datasets are presented in Table~\ref{tab:datasets_stats}.

\begin{table}[ht]
\centering
\caption{Statistics of single-database data.  
\label{tab:datasets_stats}}
\vspace{-2mm}
\begin{small}
\begin{tabular}{lccc}
\toprule %
\textbf{Dataset} & \textbf{Train} & \textbf{Dev} & \textbf{Test}   \\
\midrule %
GeoQuery & %
536 & 159 & 182 \\
IMDB & 103 & 9 & 17 \\
Yelp & 104 & 11 & 10 \\
Academic & 142 & 18 & 15\\
\bottomrule
\end{tabular}
\end{small}
\vspace{-4mm}
\end{table} 

\section{Fine-tuning Details \label{apn:finetune_params}}

For fine-tuning on single-database data, we use official implementations of all the models.\footnote{\href{https://github.com/ElementAI/picard}{github.com/ElementAI/picard}; \href{https://github.com/salesforce/TabularSemanticParsing}{github.com/salesforce/TabularSemanticParsing}; \href{https://github.com/yandex-research/sparqling-queries}{github.com/yandex-research/sparqling-queries};}
We experimented with different training strategies: initializing from released checkpoints of the models trained on Spider and training from scratch (in this case, models contain transformers pre-trained on textual data). We choose the best approach for each model and refer to it as fine-tuning.

We fine-tune T5-3B pre-trained on textual data \citep{raffel-etal-2020-t5} for 300 epochs on one NVIDIA A100 GPU with the same parameters as \citet{scholak-etal-2021-picard} used:  Adafactor optimizer \citep{adafactor2018} with learning rate 1e-4 and batch size 625. 

The released checkpoint of the BRIDGE model was trained on data that includes question splits of single-database data that we consider. We re-train this model on Spider-only data to evaluate it on query splits of single-database datasets. We use the same training parameters as \citet{lin-etal-2020-bridging} used: Adam optimizer \citep{adam2014} with the same scheduler (L-inv learning rate decay) and batch size 32. We choose the best checkpoint on the development set as \citet{lin-etal-2020-bridging} did in their work (execution accuracy and test-suite accuracy of our and authors' checkpoints are the same on Spider dev and test).
We use the same training procedure for fine-tuning on single-database data but create training data from both Spider train set and the train set of a particular dataset.

For fine-tuning SQ-QDMR on GeoQuery, we use the corresponding part of the Break dataset \citep{wolfson-etal-2020-break} and generate the 366 train groundings using the automatic annotation model of \citet{saparina-osokin-2021-sparqling}. We cannot fine-tune on the Academic dataset because its database is large and preprocessing of \citet{saparina-osokin-2021-sparqling} failed. QDMR forms for other datasets were not provided in the Break dataset, so we could not fine-tune on them.
For fine-tuning on GeoQuery, we start with the released checkpoint and parameters saved on 73000 iterations of Spider training and continue up to 81000 iterations with GeoQuery train data. We also use the same parameters as \citet{saparina-osokin-2021-sparqling} used: the optimizer is Adam \citep{adam2014} with polynomial decay scheduler used by \citep{wang-etal-2020-rat}, the batch
size is 24.

We use 1 NVIDIA A100 GPU for training T5-3B, 1 NVIDIA V100 GPU for BRIDGE and 4 NVIDIA V100 GPUs for SQ-QDMR.

\section{Execution Accuracy of Search on Single-Database Data \label{apn:others-exec}}

Table~\ref{tab:others-exec} shows execution accuracy of search on different single-database datasets. We consider two types of models: trained only on Spider data and fine-tuned on single-database data. Comparing with Table~\ref{tab:geo-testsuite}, the figures are higher because many false-positive queries pass one test (the execution accuracy metric) and do not pass the test suite. For this reason, the results of searching with the test-suite criterion are lower in terms of execution accuracy: if no tested query satisfies the test-suite criterion, the system defaults to the result of the greedy decoding, which may fail one test, while the one-test criterion would select a false positive.
\newpage
    
\begin{table}[ht!]
\centering
\caption{Different search criteria (execution, output column match, 1 test and test suite) on top of pre-trained models on the query test splits of different datasets with  \textbf{execution accuracy}. \label{tab:others-exec}}
\begin{center}
\begin{small}
\vskip -2mm
\begin{tabular}{@{}c@{\;\;}l@{}c@{\;\;}c@{\;\;}c@{\;\;}c@{\;\;}c@{}}
\toprule
\multirow{2}{*}{\shortstack{\textbf{Dataset} \\ \textbf{(test size)}}} & \multirow{2}{*}{\textbf{Model}} & \multirow{2}{*}{\textbf{Greedy}} & \multirow{2}{*}{\textbf{Exec}} & \multirow{2}{*}{\textbf{Cols}} & \multirow{2}{*}{\textbf{1 test}} & \multirow{2}{*}{\shortstack{\textbf{Test} \\ \textbf{Suite}}} \\
 &  & &  &  &  &  \\
\midrule
\multirow{6}{*}{\shortstack{GeoQuery \\ (182)}} 
& T5-3B  & 56.6 &   59.3 & 69.8 & 80.2 & 74.7\\ 
& + fine-tune & 69.8 & 76.4 & 88.5 & 97.8 & 97.3 \\[0.7ex]
& BRIDGE &  57.6  & 51.1 & 63.7 & 86.8 & 74.7 \\
& + fine-tune &  71.4 & 72.5 & 86.3 & 96.7 & 93.4 \\[0.7ex]
& SQ-QDMR &  40.7 & 40.7 & 45.6 & 92.3 & 75.8 \\
&  + fine-tune &  61.5  & 61.5 & 64.8 & 90.7  & 84.1  \\
\midrule
\multirow{5}{*}{\shortstack{IMDB \\ (17)}}
& T5-3B  & 5.9 &   17.6 & 17.6 & 35.3 & 35.3 \\ 
& + fine-tune & 52.9 & 52.9  & 52.9 & 58.8 & 52.9  \\[0.7ex]
& BRIDGE & 17.6  & 17.6 & 23.5 & 35.3 &  23.5\\
& + fine-tune &  52.9 & 52.9 & 52.9& 58.8 &  52.9\\[0.7ex]
& SQ-QDMR &  11.8 & 11.8 & 11.8 & 41.2 &  35.3 \\
\midrule
\multirow{5}{*}{\shortstack{Yelp \\ (10)}} 
& T5-3B   &  30 & 60 & 50 & 10 & 30 \\ 
& + fine-tune & 40  & 50 & 40 & 10 & 70 \\[0.7ex]
& BRIDGE & 10 & 30 & 30 & 90 & 30 \\
& + fine-tune &  40 & 80 & 80 & 100 &  70\\[0.7ex]
& SQ-QDMR &  40 & 40 & 40 & 80 & 40 \\
\midrule
\multirow{5}{*}{\shortstack{Academic \\ (15)}} 
& T5-3B   &   6.7 & 13.3 & 26.7 & - %
& 26.7 \\ 
& + fine-tune &  53.3 & 53.3 & 53.3 & - %
& 73.3 \\[0.7ex]
& BRIDGE &  6.7 & 6.7 & 6.7 & - & 20 \\
& + fine-tune & 33 & 40 & 40 & - & 80  \\[0.7ex]
& SQ-QDMR & 5.6  & 5.6 & 11.1 & - & 55.6 \\
\bottomrule
\end{tabular}
\end{small}
\end{center}
\vspace{-6mm}
\end{table}

\end{document}